\documentclass[10pt]{article} 

\usepackage[preprint]{tmlr}

\usepackage{amsmath}
\usepackage{amssymb}
\usepackage{amsthm}
\usepackage{booktabs}
\usepackage{multirow}
\usepackage{graphicx}
\usepackage{xcolor}
\usepackage{tikz}
\usetikzlibrary{arrows.meta,positioning,fit,backgrounds,calc}
\usepackage{algorithm}
\usepackage{algpseudocode}
\usepackage{url}

\usepackage[pagebackref,breaklinks,colorlinks]{hyperref}
\hypersetup{
    colorlinks=true, 
    linkcolor=blue,  
    filecolor=magenta, 
    urlcolor=blue,    
    citecolor=blue,       
}

\usepackage[capitalise,noabbrev]{cleveref}

\definecolor{hbmcol}{HTML}{C2410C}   
\definecolor{regcol}{HTML}{0E7490}   
\definecolor{predcol}{HTML}{6D28D9}  

\newcommand{\tX}{\mathsf{X}}
\newcommand{\tY}{\mathsf{Y}}
\newcommand{\tZ}{\mathsf{Z}}
\newcommand{\mW}{\mathbf{W}}
\newcommand{\mH}{\mathbf{H}}
\newcommand{\vs}{\mathbf{s}}
\newcommand{\Aop}{\mathcal{A}}          
\newcommand{\Qio}{Q}                    
\newcommand{\dwconv}{\mathbin{\ast_{\mathrm{dw}}}}

\theoremstyle{plain}
\newtheorem{proposition}{Proposition}

\title{Fast and Memory-Efficient Wavelet Convolutions\\ via I/O-Aware Reformulation}

\author{%
    \name Amit Aflalo \email aflaloa@post.bgu.ac.il \\
    \name Shahaf E. Finder \email finders@post.bgu.ac.il \\
    \name Roy Amoyal \email amoyalr@post.bgu.ac.il \\
    \name Eran Treister \email erant@bgu.ac.il \\
    \name Oren Freifeld \email orenfr@bgu.ac.il \\
    \addr The Stein Faculty of Computer and Information Science, Ben-Gurion University of the Negev, Israel
}

\begin{document}

\maketitle

\begin{abstract}
Wavelet convolution (WTConv) has emerged as an increasingly popular drop-in replacement for standard convolutions, expanding a network's receptive field exponentially with the number of decomposition levels while keeping the parameter count linear. However, its reference implementation is severely memory-bound due to excessive data movement through high-bandwidth memory (HBM). We develop an I/O model of WTConv to characterize this bottleneck and use it to guide three algebraic reformulations: (1) recomputing the inexpensive Haar analysis butterfly on chip, (2) collapsing the multi-level synthesis cascade into a single closed-form pass indexed by output-coordinate bits, and (3) folding learned per-channel scales into the convolution weights. Together, these reformulations enable an I/O-aware fused implementation that substantially reduces HBM traffic. We evaluate the WTConvNeXt configuration across decomposition levels and a broad range of tensor shapes. Despite performing comparable arithmetic, the reference WTConv is substantially slower than the depthwise convolution it replaces. Our reformulation reduces modeled HBM traffic by approximately $2.55\times$, yielding up to a $4.35\times$ training speedup over the reference while roughly halving peak memory usage. Thus, our reformulation preserves the benefits of WTConv while substantially reducing its execution time and memory footprint, removing the systems overhead that previously limited its practical efficiency. Source code is available in the official WTConv repository under \href{https://github.com/BGU-CS-VIL/WTConv/tree/main/fast_wtconv}{fast\_wtconv}.

\end{abstract}

\section{Introduction}

Large receptive fields are important for modern convolutional networks, but obtaining them with conventional convolutions is expensive. Stacking small kernels expands the theoretical receptive field only linearly with depth, while directly increasing kernel size incurs parameter and arithmetic costs proportional to the kernel area \citep{luo2016erf,ding2022replknet,liu2023slak}. 
Fortunately, the widely-used WTConv \citep{finder2024wavelet} offers an appealing alternative: it applies small depthwise convolutions across the progressively downsampled levels of a wavelet decomposition, so the receptive field grows exponentially with the number of levels while the parameter count grows only linearly. This makes WTConv a drop-in replacement for large depthwise convolutions. WTConv has also produced accuracy and robustness gains in architectures including ConvNeXt and MobileNetV2 \citep{liu2022convnext,sandler2018mobilenetv2}.

Yet this advantage does not translate into execution speed. In the original WTConv paper, WTConvNeXt uses WTConv with a $5\times5$ kernel ($k{=}5$) as a replacement for the $7\times7$ depthwise convolution in a ConvNeXt block. In this setting, the reference implementation is \emph{slower than the convolution it replaces}: over a full training step, it trails the depthwise $7\times7$ baseline by $2.46$--$3.42\times$ in \texttt{fp32} and $1.53$--$2.20\times$ in \texttt{fp16}; at inference, the corresponding gaps are $3.79$--$5.28\times$ and $2.05$--$2.70\times$. These gaps cannot be explained by arithmetic alone. For an input containing $N=B\cdot C\cdot H\cdot W$ elements, WTConv with $k{=}5$ performs $58N$--$69N$ multiply--accumulates as the number of decomposition levels increases from $L{=}1$ to $L{=}5$, compared with $49N$ for the $7\times7$ depthwise baseline. This increase in arithmetic is far smaller than the observed latency gap. Thus, despite its favorable parameter scaling, the reference operator incurs a substantial wall-clock penalty. Closing this performance gap is the primary objective of this work.

We show that the discrepancy arises because FLOPs are the wrong cost model for this operator. The reference WTConv implementation has an arithmetic intensity of only $\approx1.63$~FLOP/byte in \texttt{fp32}, placing it deep in the memory-bound regime on modern GPUs. Its execution is therefore dominated not by the cost of the Haar transform or the depthwise convolutions themselves, but by repeatedly materializing intermediate wavelet coefficients and reconstructions in high-bandwidth memory (HBM). Under a tensor-materialization I/O model, a forward evaluation incurs $17.75N$--$21.32N$ element reads and writes to global memory for an input containing $N$ elements.

This observation suggests a different optimization target: rather than reducing arithmetic, we reformulate WTConv so that mathematical intermediates need not become memory-resident intermediates. We exploit three properties of the operator. First, Haar analysis uses only signed additions and fixed power-of-two scaling and can be recomputed on chip inside the depthwise convolution. Second, linearity of Haar synthesis allows the entire $L$-level reconstruction recursion to be written as a single closed-form sum whose signs and coefficient addresses are determined by bits of the output coordinate. Third, the learned per-channel scales can be folded into the convolution weights. Together, these transformations preserve the same mathematical operator while eliminating the large intermediate tensors responsible for the majority of its data movement. The resulting formulation reduces modeled HBM traffic by $2.54$--$2.56\times$ for $L=1,\ldots,5$.

We implement this formulation in CUDA and evaluate both inference and full training steps over a broad sweep of tensor shapes, decomposition levels, and precisions. For a full training step, the implementation is $3.71$--$4.35\times$ faster than the reference in \texttt{fp32} and $2.68$--$3.09\times$ faster in \texttt{fp16}, while reducing peak memory by a factor of $1.83$--$2.31$. More importantly, the systems reformulation reverses the practical comparison that motivates the work: over a training step the optimized WTConv at $k{=}5$ is $1.27$--$1.50\times$ faster in \texttt{fp32} and $1.40$--$1.76\times$ faster in \texttt{fp16} than the depthwise $7\times7$ convolution it is proposed to replace, at every decomposition level tested. Training is the regime in which the comparison matters most, since it is where the materialized intermediates are both written and re-traversed; at inference the reversal is complete in \texttt{fp16} but not in \texttt{fp32}, where the fused layer remains within $3$--$17\%$ of the depthwise baseline rather than ahead of it (\autoref{sec:results-latency-inference}).

\paragraph{Contributions.}
\begin{itemize}\itemsep2pt

\item \textbf{An I/O cost model for WTConv.}
We derive an element-level accounting of HBM traffic for the reference implementation,
$
\Qio_{\mathrm{ref}}
=7N+\tfrac{43}{3}N(1-4^{-L}),
$
which is independent of the kernel size, and use a roofline analysis to show that WTConv's arithmetic intensity is a factor of roughly $31$ below the compute-saturation ridge point in \texttt{fp32} at $k{=}5$. This identifies data movement, rather than arithmetic, as its dominant cost.

\item \textbf{An algebraically exact, I/O-aware reformulation.}
We combine register-resident Haar analysis, a closed-form bit-indexed synthesis over all decomposition levels, and scale folding to eliminate unnecessary HBM-resident intermediates. The resulting formulation reduces predicted forward-pass traffic by $2.54$--$2.56\times$ across $L=1,\ldots,5$ while preserving the WTConv operator up to floating-point evaluation order.

\item \textbf{End-to-end empirical validation.}
Across all measured configurations, our CUDA implementation substantially reduces training and inference latency and peak memory relative to the reference WTConv implementation, and over a training step outperforms the depthwise $7\times7$ convolution WTConv is proposed to replace. We additionally verify numerical agreement for the forward output and all parameter gradients.

\end{itemize}

\section{Background and Related Work}
\label{sec:background}

\subsection{The WTConv operator}

Let $\tX\in\mathbb{R}^{B\times C\times H\times W}$ denote the input, and let $N=B\cdot C\cdot H\cdot W$ be its element count, the unit in which we quote every I/O cost. Let $\dwconv$ denote depthwise convolution (one $k\times k$ kernel per channel, zero-padded to preserve resolution) and let $\odot$ denote per-channel scaling.

Let $\Aop$ denote one level of Haar analysis, which splits each channel into four half-resolution subbands $(\mathrm{LL},\mathrm{LH},\mathrm{HL},\mathrm{HH})$, and let $\Aop^{\!\top}$ denote its synthesis, which inverts it exactly (\autoref{prop:selfinverse}). For a tensor carrying a subband axis, let $[\,\cdot\,]_{\mathrm{LL}}$ denote the low-pass band, $[\,\cdot\,]_{\mathrm{H}}$ the three high-pass bands, and $\Vert$ their concatenation. WTConv preserves the channel count. It learns a $k\times k$ kernel and a per-channel scale at each level, $(\mW^{(\ell)},\vs^{(\ell)})$ for $\ell=1,\dots,L$, together with a full-resolution base convolution $(\mW^{(0)},b,\vs^{(0)})$, giving $\mathcal{O}(Lk^2C)$ parameters in total. \autoref{alg:wtconv} states the operator as the reference implements it.

\begin{algorithm}[!ht]
\caption{WTConv, reference implementation \citep{finder2024wavelet}}
\label{alg:wtconv}
\begin{algorithmic}[1]
\Require input $\tX$, levels $L$, parameters $\{\mW^{(\ell)},\vs^{(\ell)}\}_{\ell=0}^{L}$, bias $b$
\State $\tX^{(0)} \gets \tX$ \Comment{decomposition carrier}
\For{$\ell = 1,\dots,L$} \Comment{downward: decompose and filter}
  \State $\tY^{(\ell)} \gets \Aop\,\tX^{(\ell-1)}$ \Comment{Haar analysis}
  \State $\tZ^{(\ell)} \gets \vs^{(\ell)}\odot\bigl(\tY^{(\ell)}\dwconv\mW^{(\ell)}\bigr)$ \Comment{filter, then scale}
  \State $\tX^{(\ell)} \gets \bigl[\tY^{(\ell)}\bigr]_{\mathrm{LL}}$ \Comment{carry the \emph{raw}, unfiltered low-pass band}
\EndFor
\State $R^{(L+1)} \gets 0$
\For{$\ell = L,\dots,1$} \Comment{upward: reconstruct and accumulate}
  \State $R^{(\ell)} \gets \Aop^{\!\top}\Bigl(\bigl(\bigl[\tZ^{(\ell)}\bigr]_{\mathrm{LL}} + R^{(\ell+1)}\bigr)\ \Big\Vert\ \bigl[\tZ^{(\ell)}\bigr]_{\mathrm{H}}\Bigr)$
\EndFor
\State \Return $\vs^{(0)}\odot\bigl(\tX\dwconv\mW^{(0)} + b\bigr) + R^{(1)}$ \Comment{base path $+$ reconstruction}
\end{algorithmic}
\end{algorithm}

\subsection{Haar transform differentiation}
\label{sec:haar-backprop}

For each non-overlapping $2\times2$ block, the normalized Haar transform maps
four input samples to four subband coefficients using the following matrix.

\begin{proposition}[Symmetry and self-inversion]
\label{prop:selfinverse}
$\mH=\tfrac12\left[\begin{smallmatrix}1&1&1&1\\1&1&-1&-1\\1&-1&1&-1\\1&-1&-1&1\end{smallmatrix}\right]$
satisfies $\mH^{\!\top}=\mH$ and $\mH^2=\mathbf{I}$. Therefore,
$\mH^{-1}=\mH^{\!\top}=\mH$.
\end{proposition}

The full analysis operator $\Aop$ applies $\mH$ independently to every block,
and synthesis applies $\Aop^{\!\top}$. For $\mathbf{y}=\mH\mathbf{x}$, the
chain rule gives
\begin{equation}
\frac{\partial\mathcal{L}}{\partial\mathbf{x}}
= \mH^{\!\top}\frac{\partial\mathcal{L}}{\partial\mathbf{y}}
= \mH\frac{\partial\mathcal{L}}{\partial\mathbf{y}}.
\end{equation}
Thus, backward through analysis applies synthesis, and backward through
synthesis applies analysis. The Haar step needs no separate derivative;
convolution and scale gradients are computed as usual.

\subsection{The roofline}
\label{roofline}

The roofline model \citep{williams2009roofline} bounds attainable throughput by $\min(\pi,\, \beta I)$, where $\pi$ is peak compute, $\beta$ peak bandwidth, and $I$ the arithmetic intensity in FLOP/byte. The ridge point $I^\star=\pi/\beta$ separates memory-bound from compute-bound operation. For the RTX~A6000 used throughout, $\beta=768$~GB/s and $\pi\approx38.7$~TFLOP/s in \texttt{fp32}, so $I^\star\approx50$~FLOP/byte. Any operator with $I\ll I^\star$ is bandwidth-limited: its runtime is governed by the bytes it moves rather than by how the arithmetic is scheduled. The bound is an upper bound on attainable throughput and not an equality, so this identifies traffic as the quantity to minimize without implying that runtime is proportional to it. The practical content of this paper is that WTConv is deeply in that regime, so minimizing traffic is not merely one optimization among many; it is the dominant optimization target.

\subsection{Kernel fusion and memory-bound operators} It is well established that data movement, rather than arithmetic, is often the dominant cost in deep learning workloads \citep{ivanov2021datamovement,wahib2014kernelfusion}. FlashAttention \citep{dao2022flashattention,dao2024flashattention2} provides the canonical example: by exactly reformulating the computation so that intermediate results remain on chip, it substantially reduces the practical cost of attention. Our work applies the same principle to a different operator, but exploits additional algebraic structure: the fused transform uses only signed additions and fixed power-of-two scaling, and is symmetric and self-inverse. These properties make recomputation inexpensive and allow analysis and synthesis to exchange roles during backpropagation. General-purpose compilers \citep{ragankelley2013halide,chen2018tvm,sabne2020xla,ansel2024pytorch2} can automate fusion across elementwise and reduction chains, but the reformulation developed in \autoref{sec:synthesis} (which collapses an $L$-step recursion into a closed-form, bit-indexed sum) depends on specific properties of the Haar basis that are not available to a general-purpose compiler.

\section{WTConv is memory-bound}
\label{sec:memorybound}

\subsection{I/O cost of the reference implementation}

\begin{figure}[t]
\centering
\begin{tikzpicture}[
  font=\small,
  node distance=0pt,
  hbm/.style={draw=hbmcol!80!black, fill=hbmcol!18, rounded corners=1.5pt,
              minimum height=6.5mm, minimum width=13mm, align=center, inner sep=2pt},
  reg/.style={draw=regcol!75!black, fill=regcol!18, rounded corners=1.5pt,
              minimum height=6.5mm, minimum width=13mm, align=center, inner sep=2pt},
  op/.style={draw=black!55, fill=black!5, rounded corners=1pt,
             minimum height=5mm, minimum width=9mm, align=center, inner sep=1.5pt,
             font=\scriptsize},
  flow/.style={-{Stealth[length=4pt,width=3.5pt]}, draw=black!65, line width=0.5pt},
  ghost/.style={draw=black!35, dashed, fill=none, rounded corners=1.5pt,
                minimum height=5.5mm, minimum width=12mm, align=center, inner sep=2pt,
                font=\scriptsize, text=black!55},
  lbl/.style={font=\scriptsize\itshape, text=black!60},
]

\node[lbl, font=\normalsize\itshape ,anchor=west] at (0,0.95) {(a) Baseline: every stage round-trips through HBM};

\node[hbm, anchor=west] (b0) at (0,0) {$\tX$\\[-1pt]\scriptsize $N_\ell$};
\node[op,  right=4.5mm of b0]   (bop1) {\texttt{conv2d}\\[-2pt]\scriptsize WT};
\node[hbm, right=4.5mm of bop1] (b1)   {$\tY$\\[-1pt]\scriptsize $N_\ell$};
\node[op,  right=4.5mm of b1]   (bop2) {\texttt{conv2d}\\[-2pt]\scriptsize dw};
\node[hbm, right=4.5mm of bop2] (b2)   {$\tZ$\\[-1pt]\scriptsize $N_\ell$};
\node[op,  right=4.5mm of b2]   (bop3) {\texttt{mul}\\[-2pt]\scriptsize scale};
\node[hbm, right=4.5mm of bop3] (b3)   {$\tZ'$\\[-1pt]\scriptsize $N_\ell$};
\node[op,  right=4.5mm of b3]   (bop4) {\texttt{slice}\\[-2pt]\scriptsize LL};
\node[hbm, right=4.5mm of bop4] (b4)   {$\tX^{(\ell+1)}$\\[-1pt]\scriptsize $\tfrac{1}{4}N_\ell$};

\foreach \a/\b in {b0/bop1, bop1/b1, b1/bop2, bop2/b2, b2/bop3, bop3/b3, b3/bop4, bop4/b4}
  \draw[flow] (\a) -- (\b);

\node[lbl, font=\normalsize\itshape,anchor=west] at (0,-1.35) {(b) Fused: one read, one pass --- the transform is recomputed in registers};

\node[ghost] (g1) at (3.65,-2.25) {$\tY$};
\node[ghost, right=4mm of g1] (g2) {$\tZ$};
\node[lbl, right=3mm of g2] {never written to HBM};

\node[hbm, anchor=west] (f0) at (0,-3.25) {$\tX$\\[-1pt]\scriptsize $N_\ell$};

\node[reg, anchor=west, minimum width=46mm, minimum height=13mm] (freg) at (2.05,-3.25)
  {\scriptsize register-resident\\[-1pt]\scriptsize
   $2{\times}2$ load $\to$ butterfly $\to$ MAC\\[-1pt]\scriptsize
   $\widetilde{\mW}=\vs\odot\mW$ folded};

\node[hbm, anchor=west] (f3) at (8.45,-2.85) {$\tZ'$\\[-1pt]\scriptsize $N_\ell$};
\node[hbm, anchor=west] (f4) at (8.45,-3.72) {$\tX^{(\ell+1)}$\\[-1pt]\scriptsize $\tfrac{1}{4}N_\ell$};

\draw[flow] (f0) -- node[lbl, above, pos=0.45] {read} (freg);
\draw[flow] (freg.east) -- node[lbl, above, pos=0.6, yshift=-0.5pt] {write} (f3.west);
\draw[flow] (freg.east) -- (f4.west);

\draw[densely dotted, draw=black!35] (g1.south) -- ++(0,-4mm);
\draw[densely dotted, draw=black!35] (g2.south) -- ++(0,-4mm);

\begin{scope}[shift={(0,-4.75)}]
  \node[hbm, minimum width=5mm, minimum height=3.6mm, anchor=west] (lg1) at (0,0) {};
  \node[anchor=west, font=\scriptsize, right=1.5mm of lg1] (lt1) {HBM-resident tensor};
  \node[reg, minimum width=5mm, minimum height=3.6mm, anchor=west, right=6mm of lt1] (lg2) {};
  \node[anchor=west, font=\scriptsize, right=1.5mm of lg2] (lt2) {on-chip (registers)};
  \node[ghost, minimum width=5mm, minimum height=3.6mm, anchor=west, right=6mm of lt2] (lg3) {};
  \node[anchor=west, font=\scriptsize, right=1.5mm of lg3] {eliminated intermediate};
\end{scope}

\end{tikzpicture}
\caption{HBM dataflow for one WTConv decomposition level $\ell$, with $N_\ell = N/4^{\ell-1}$ the number of elements entering that level. (a)~The reference implementation expresses the level as a chain of framework primitives; every stage materializes a full-size tensor in HBM. (b)~The fused formulation reads the input once, evaluates the Haar butterfly and the depthwise convolution on chip against weights that already carry the learned scale, and writes only the results. $\tY$ and $\tZ$ are never written.}
\label{fig:dataflow}
\end{figure}
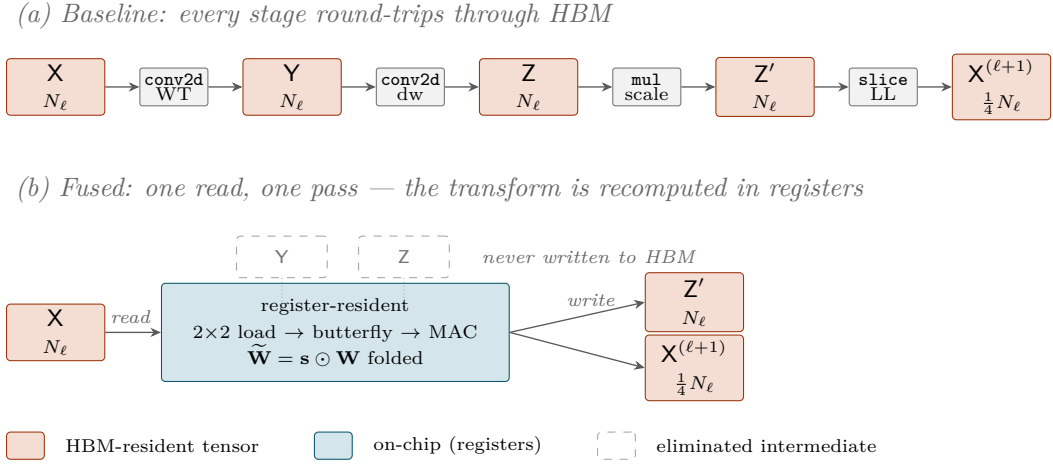

The reference implementation expresses each level as a chain of framework primitives, every one of which reads its input from HBM and writes its output back (\hyperref[fig:dataflow]{\autoref*{fig:dataflow}a}). We count elements crossing the HBM boundary: Each stage reads its input tensor once and writes its output tensor once. Weight traffic is $\mathcal{O}(Ck^2)$ and therefore negligible against $\mathcal{O}(N)$. Writing $N_\ell = N/4^{\ell-1}$ for the element count entering level $\ell$:

\begin{itemize}\itemsep1pt
\item \emph{Analysis}, per level: the Haar transform is executed as a grouped stride-2 \texttt{conv2d} against a constant $\pm\tfrac12$ filter ($2N_\ell$); the depthwise convolution over the $4C$-channel coefficient tensor ($2N_\ell$); the scale multiply ($2N_\ell$). Total $6N_\ell$.
\item \emph{Synthesis}, per level: the cross-level low-pass add ($\tfrac34 N_\ell$); the concatenation of the summed low-pass band with the three high bands ($2N_\ell$); the \texttt{conv\_transpose2d} ($2N_\ell$). Total $\tfrac{19}{4}N_\ell$.
\item \emph{Base path}: convolution ($2N$), scale multiply ($2N$), final add ($3N$). Total $7N$.
\end{itemize}

Using $\sum_{\ell=1}^{L}4^{-(\ell-1)}=\tfrac43(1-4^{-L})$,
\begin{equation}
\Qio_{\mathrm{ref}} \;=\; \underbrace{7N}_{\text{base path}} \;+\; \underbrace{\Bigl(6+\tfrac{19}{4}\Bigr)}_{\text{per-level cost}}\cdot\underbrace{\tfrac43\,N\bigl(1-4^{-L}\bigr)}_{\sum_{\ell=1}^L N_\ell}
\;=\; 7N + \tfrac{43}{3}\,N\bigl(1-4^{-L}\bigr),
\label{eq:qref}
\end{equation}
rising from $17.75N$ at $L=1$ to $21.33N$ as $L\to\infty$; that is, evaluating the layer moves between $18$ and $21$ times the input tensor's worth of data through HBM. Note that $k$ does not appear: the kernel size sets how much arithmetic each resident element receives, not how many elements cross the boundary. For a representative $B{=}8$, $C{=}64$, $H{=}W{=}256$ input, $N=33.6$M elements ($128$\,MiB in \texttt{fp32}) and the reference therefore moves roughly $2.9$\,GB per forward pass. 

\subsection{Arithmetic intensity of the reference implementation}

 The base convolution performs $k^2$ multiply--accumulates per element. At level $\ell$, the depthwise convolution over the coefficient tensor costs a further $k^2$ per element of $N_\ell$, and the analysis and synthesis each cost $4$, since the reference executes both as $2\times2$ convolutions rather than as additions. Summing over levels with the same geometric factor as before,
\begin{equation}
\mathrm{MAC} \;=\; N\bigl[k^2 + (k^2+8)\cdot\tfrac43\bigl(1-4^{-L}\bigr)\bigr].
\label{eq:mac}
\end{equation}
For $k=5$ and $L=5$ this is $69.0N$ multiply--accumulates, or $137.9N$ FLOPs, against $4\Qio_{\mathrm{ref}}=85.3N$ bytes in \texttt{fp32}:
\begin{equation}
I_{\mathrm{ref}} \;=\; \frac{137.9N}{85.3N} \;\approx\; 1.63~\text{FLOP/byte}.
\label{eq:intensity}
\end{equation}
 Against the ridge point $I^\star\approx50$ (\autoref{roofline}), WTConv's arithmetic intensity is lower by a factor of roughly $31$ in \texttt{fp32}; equivalently, it is approximately $3.2\%$ of the ridge-point intensity. Enlarging the kernel does not change this conclusion, only its margin: increasing $k$ increases the arithmetic work while leaving \autoref{eq:qref} untouched, so even at $k{=}5$ the operator remains more than an order of magnitude below the ridge point. Arithmetic is not the bottleneck: accelerating the transform itself would have little effect, whereas reducing memory traffic directly targets the dominant cost.

\section{An I/O-Aware formulation}
\label{sec:method}

To eliminate the severe memory bottleneck of the reference WTConv, we derive an I/O-aware formulation based on three exact algebraic reformulations. These reformulations leave the underlying mathematical operator unchanged, differing from the reference implementation solely in floating-point evaluation order. Together, they eliminate the unnecessary Haar-analysis and reconstruction intermediates from high-bandwidth memory (HBM) by performing multi-stage computations directly in GPU registers and shared memory across all decomposition levels.

\subsection{Fusing Haar analysis with depthwise convolution}
\label{sec:analysis}

For a $2\times2$ spatial patch $\left[\begin{smallmatrix}a & b \\ c & d\end{smallmatrix}\right]$, the normalized 2D Haar analysis transform maps four input pixels to subband coefficients $(\mathrm{LL}, \mathrm{LH}, \mathrm{HL}, \mathrm{HH})$ as follows:
\begin{equation}
\begin{aligned}
\mathrm{LL} &= \tfrac12(a+b+c+d), &\qquad \mathrm{LH} &= \tfrac12(a+b-c-d),\\
\mathrm{HL} &= \tfrac12(a-b+c-d), &\qquad \mathrm{HH} &= \tfrac12(a-b-c+d).
\end{aligned}
\label{eq:butterfly}
\end{equation}
Evaluating \autoref{eq:butterfly} requires only signed additions and fixed scaling by $\tfrac12$. In contrast, the reference implementation materializes the entire coefficient tensor $\tY^{(\ell)}$ in HBM via grouped convolutions with $\pm\tfrac12$ filters. Each level runs Haar analysis, depthwise convolution, and learned scaling as separate passes. Each pass reads and writes an $N_\ell$-element tensor, moving $6N_\ell$ elements in total. Our fused approach computes the Haar coefficients inside the depthwise convolution and folds the scale into its weights (\autoref{sec:scale_folding}; \hyperref[fig:dataflow]{\autoref*{fig:dataflow}b}). Because WTConv reaches only $1.63$~FLOP/byte (\autoref{sec:memorybound}), recomputing the coefficients is much cheaper than materializing them in HBM.

Naive fusion would recompute the transform for every convolution tap and load each input pixel $k^2$ times. Instead, each thread block transforms the $2\times2$ blocks for one output tile, including a coefficient-space halo of radius $R=(k-1)/2$, and stages the four subbands in shared memory. All convolution taps reuse the transformed tile, so each $2\times2$ input block is transformed only once per output tile. Except at the deepest level, the fused approach reads $N_\ell$ input elements, writes $N_\ell$ filtered coefficients for synthesis, and writes $\tfrac14N_\ell$ unfiltered low-pass coefficients for the next level. It therefore moves $\tfrac94N_\ell$ elements, versus $6N_\ell$ for the reference. The deepest level has no successor and moves $2N_L$.

\subsection{Collapsing the synthesis cascade}
\label{sec:synthesis}

The reference implementation executes reconstruction as an $L$-step sequential loop, where each level reads the lower-level reconstruction from HBM, adds it to its low-pass subband, and applies transposed convolution. This imposes $L$ strict sequential dependencies and incurs an I/O cost of $\tfrac{19}{4}N_\ell$ per level.

Because Haar synthesis is linear, the low-pass carrier acts as a linear accumulator across levels. As a result, the multi-level synthesis cascade can be collapsed into a single closed-form pass across all $L$ levels. Haar synthesis at level $\ell$ expands each subband coefficient over a $2^\ell \times 2^\ell$ pixel block using a fixed sign pattern determined by the spatial location of the output pixel within that block.

\begin{proposition}[Bit-indexed single-pass synthesis]
\label{prop:bitindexed}
Let $(y,x)$ denote an output pixel coordinate, and let level $\ell \in \{1,\dots,L\}$ have subband coefficients $(c^{(\ell)}_{\mathrm{LL}},c^{(\ell)}_{\mathrm{LH}},c^{(\ell)}_{\mathrm{HL}},c^{(\ell)}_{\mathrm{HH}})$ at coarse grid location $(\lfloor y / 2^\ell \rfloor, \lfloor x / 2^\ell \rfloor)$. Define the coordinate parity signs at level $\ell$ as follows:
\begin{equation}
s_y^{(\ell)} = (-1)^{\lfloor y / 2^{\ell-1} \rfloor \bmod 2}, \qquad
s_x^{(\ell)} = (-1)^{\lfloor x / 2^{\ell-1} \rfloor \bmod 2}.
\end{equation}
The reconstructed pixel value $R_{y,x}$ across all $L$ levels is given by:
\begin{equation}
R_{y,x} = \sum_{\ell=1}^{L} 2^{-\ell} \Bigl(
  c^{(\ell)}_{\mathrm{LL}} +
  s_y^{(\ell)} c^{(\ell)}_{\mathrm{LH}} +
  s_x^{(\ell)} c^{(\ell)}_{\mathrm{HL}} +
  s_y^{(\ell)} s_x^{(\ell)} c^{(\ell)}_{\mathrm{HH}}
\Bigr).
\label{eq:synthesis_closed_form}
\end{equation}
\end{proposition}

A derivation of the coordinate signs and normalization in
\autoref{eq:synthesis_closed_form} is given in \autoref{app:bitindexed-proof}.

By evaluating \autoref{eq:synthesis_closed_form} in a single pass, a GPU thread computing pixel $(y,x)$ iterates from level $L$ down to $1$, addressing coarse coefficients via coordinate bit-shifts and obtaining signs through bitwise parity checks. This eliminates all intermediate low-pass tensors and sequential kernel dependencies. Furthermore, the base-path convolution output is folded directly into the final output store, eliminating a separate full-resolution tensor addition.

\subsection{Folding the learned scale}
\label{sec:scale_folding}

The reference implementation applies learned per-channel scaling $s_c$ after convolution ($\tZ_c = s_c (\tX \dwconv \mW)_c$) as a standalone elementwise pass. At an arithmetic intensity of $\tfrac18$~FLOP/byte in \texttt{fp32} (one multiply per element read and written), this standalone pass is strongly memory-bound and adds an avoidable read--write traversal of the tensor.

By the linearity of convolution, the scale factor is folded into the depthwise weights and, on the base path, into its bias:
\begin{equation}
\widetilde{\mW}_c = s_c \odot \mW_c, \qquad \widetilde{b}_c = s_c\, b_c,
\label{eq:fold}
\end{equation}
so that $\tZ = \tX \dwconv \widetilde{\mW} + \widetilde{b}$ is computed in a single pass with zero extra runtime cost.

During backpropagation, exact parameter gradients with respect to the unscaled weight $\mW_c$, bias $b_c$, and scale $s_c$ are efficiently recovered as follows:
\begin{equation}
\frac{\partial\mathcal{L}}{\partial\mW_c} = s_c \, \frac{\partial\mathcal{L}}{\partial\widetilde{\mW}_c}, \qquad
\frac{\partial\mathcal{L}}{\partial b_c} = s_c \, \frac{\partial\mathcal{L}}{\partial\widetilde{b}_c}, \qquad
\frac{\partial\mathcal{L}}{\partial s_c} = \left\langle \frac{\partial\mathcal{L}}{\partial\widetilde{\mW}_c}, \, \mW_c \right\rangle + \frac{\partial\mathcal{L}}{\partial\widetilde{b}_c}\, b_c.
\label{eq:foldgrad}
\end{equation}
$s_c$ enters both folds, so its gradient collects a term from each.

\subsection{The resulting I/O cost}
\label{sec:prediction}

Collecting the reformulations across all levels $1,\dots,L$, with $N_\ell = N/4^{\ell-1}$:
\begin{itemize}\itemsep1pt
\item \emph{Base path}: the folded vendor convolution reads $N$ and writes $N$ ($2N$).
\item \emph{Analysis \& Depthwise pass}, levels $1\le\ell\le L$: reads $N_\ell$, writes $N_\ell$ convolved coefficients and $\tfrac14N_\ell$ raw low-pass coefficients ($\tfrac94N_\ell$ each, less the $\tfrac14N_L$ the deepest level does not write).
\item \emph{Single-pass Synthesis}: reads all subband coefficients $\sum_{\ell=1}^L N_\ell$ and base-path output $N$, writing the final output $N$ ($\sum_{\ell=1}^L N_\ell + 2N$).
\end{itemize}
Using $\sum_{\ell=1}^{L}N_\ell=\tfrac43N\bigl(1-4^{-L}\bigr)$, the total memory traffic is:
\begin{equation}
\Qio_{\mathrm{fused}} \;=\; 4N \;+\; \tfrac{13}{3}\,N\bigl(1-4^{-L}\bigr) \;-\; N\,4^{-L}.
\label{eq:qfused}
\end{equation}
Comparing this to the reference memory traffic $\Qio_{\mathrm{ref}}$ (\autoref{eq:qref}), the theoretical reduction ratio in HBM data movement is:
\begin{equation}
\frac{\Qio_{\mathrm{ref}}}{\Qio_{\mathrm{fused}}}
= \frac{7 + \tfrac{43}{3}\bigl(1-4^{-L}\bigr)}{4 + \tfrac{13}{3}\bigl(1-4^{-L}\bigr) - 4^{-L}}
\;\xrightarrow[L\to\infty]{}\; \frac{64}{25} = 2.56.
\label{eq:qratio}
\end{equation}
As $L \to \infty$, the fused formulation achieves up to a $2.56\times$ reduction in memory traffic over the reference implementation. \autoref{tab:traffic} summarizes the exact element counts and traffic reduction ratios for decomposition levels $L{=}1,\dots,5$. Like \autoref{eq:qref}, neither count depends on $k$, so the traffic argument of this section is stated once and holds at every kernel size; \autoref{sec:results-kernel-size} verifies this empirically at $k{=}3$.

We test both predictions using hardware performance counters in \autoref{app:traffic}. Under the modeled access pattern, the fused and reference traffic agree with their predictions within $1\%$ and $4\%$, respectively. Across tensor shapes, fused traffic remains within $5\%$ of the prediction, whereas the reference can exceed it by up to $39\%$ when cuDNN selects kernels with additional DRAM accesses. Thus, $\Qio_{\mathrm{ref}}$ should be interpreted as algorithmic traffic and, under such dispatches, as a lower bound.

\begin{table}[h]
\centering
\caption{Elements crossing the HBM boundary per forward evaluation, in units of $N=B\cdot C\cdot H\cdot W$, from \autoref{eq:qref}, \autoref{eq:qfused}, and \autoref{eq:qratio}. Both counts are independent of the kernel size $k$. The last row is a ratio of \emph{traffic}, not of runtime; see \autoref{roofline}.}
\label{tab:traffic}
\begin{tabular}{lccccc}
\toprule
& \multicolumn{5}{c}{Decomposition levels $L$}\\
\cmidrule(lr){2-6}
 & 1 & 2 & 3 & 4 & 5 \\
\midrule
Reference, $\Qio_{\mathrm{ref}}$ & $17.75N$ & $20.44N$ & $21.11N$ & $21.28N$ & $21.32N$ \\
Fused, $\Qio_{\mathrm{fused}}$   & $7.00N$  & $8.00N$  & $8.25N$  & $8.31N$  & $8.33N$ \\
\addlinespace[2pt]
Traffic reduction factor & $2.54\times$ & $2.55\times$ & $2.56\times$ & $2.56\times$ & $2.56\times$ \\
\bottomrule
\end{tabular}
\end{table}

\section{Results}
\label{sec:results}

\subsection{Setup}
\label{sec:setup}

All measurements are single-layer microbenchmarks on an RTX~A6000, except the end-to-end networks in \autoref{sec:results-endtoend} and the cross-hardware validation in \autoref{sec:results-hardware}, which repeats the layer sweep on a second, architecturally different GPU. We sweep $C\in\{32,64,128\}$, $H=W\in\{128,256,512\}$ at $B{=}8$, and $L\in\{1,\dots,5\}$ in \texttt{fp32} and \texttt{fp16}. Timings are averaged over 50 iterations following 20 warm-up iterations.

\paragraph{Baselines.} WTConv (reference and fused) runs at $k{=}5$ throughout
\autoref{sec:results-latency-train} and \autoref{sec:results-memory}; \autoref{sec:results-kernel-size} repeats the sweep at $k{=}3$, and \autoref{sec:results-hardware} repeats it at $k{=}5$ on a second GPU. We evaluate plain depthwise convolutions under two protocols: \emph{matched} ($k{=}5$) to isolate wavelet overhead, and \emph{drop-in} ($k{=}7$) representing the standard ConvNeXt replacement \citep{liu2022convnext}.

\paragraph{Reporting.} Table values are the geometric mean over the $(C,H)$ sweep of the per-configuration ratio between the WTConv reference and the evaluated method. The reference is always $1.00\times$. Latency is reported as speedup (reference/method; higher is better) and memory as footprint fraction (method/reference; lower is better). \autoref{app:extra} verifies numerical agreement.
\subsection{Layer latency}
\label{sec:results-latency-train}

\paragraph{Training step.} \autoref{tab:latency-train} reports full training-step
(forward and backward) latency. The reference operator is slower than the convolution it
replaces: the depthwise $7\times7$ convolution of a ConvNeXt block completes a
training step $2.46$--$3.42\times$ faster in \texttt{fp32} and
$1.53$--$2.20\times$ faster in \texttt{fp16}; at matched kernel size the margin
widens to $4.89$--$6.79\times$ (\texttt{fp32}, depthwise $5\times5$).

\begin{table}[htpb]
\centering
\caption{Latency of a full training step (forward and backward) relative to the reference WTConv ($1.00\times$; higher is faster).}
\label{tab:latency-train}
\begin{tabular}{llccccc}
\toprule
\multirow{2}{*}{Method} & \multirow{2}{*}{Precision} & \multicolumn{5}{c}{Decomposition levels $L$} \\
\cmidrule(lr){3-7}
 & & 1 & 2 & 3 & 4 & 5 \\
\midrule
Depthwise conv, $k{=}5$ & \texttt{fp32} & 4.89$\times$ & 6.21$\times$ & 6.60$\times$ & 6.74$\times$ & 6.79$\times$ \\
 & \texttt{fp16} & 6.28$\times$ & 8.12$\times$ & 8.67$\times$ & 8.90$\times$ & 9.07$\times$ \\
\addlinespace[2pt]
Depthwise conv, $k{=}7$ & \texttt{fp32} & 2.46$\times$ & 3.13$\times$ & 3.33$\times$ & 3.40$\times$ & 3.42$\times$ \\
 & \texttt{fp16} & 1.53$\times$ & 1.97$\times$ & 2.11$\times$ & 2.16$\times$ & 2.20$\times$ \\
\addlinespace[2pt]
WTConv, reference & \texttt{fp32} & 1.00$\times$ & 1.00$\times$ & 1.00$\times$ & 1.00$\times$ & 1.00$\times$ \\
 & \texttt{fp16} & 1.00$\times$ & 1.00$\times$ & 1.00$\times$ & 1.00$\times$ & 1.00$\times$ \\
\addlinespace[2pt]
WTConv, \textbf{fused} & \texttt{fp32} & 3.71$\times$ & 4.23$\times$ & 4.34$\times$ & 4.35$\times$ & 4.33$\times$ \\
 & \texttt{fp16} & 2.68$\times$ & 3.02$\times$ & 3.08$\times$ & 3.09$\times$ & 3.08$\times$ \\
\addlinespace[2pt]
\bottomrule
\end{tabular}
\end{table}

Fusion reverses this. The fused layer is $3.71$--$4.35\times$ faster than the
reference in \texttt{fp32} and $2.68$--$3.09\times$ in \texttt{fp16}. The ratio
grows with $L$ and then flattens, because deeper decompositions give the reference
more intermediates to materialize while \autoref{eq:qfused} is nearly flat in $L$;
the change from $L{=}4$ to $L{=}5$ is below $1\%$ in both precisions, as
\autoref{eq:qratio} predicts. At every level and in both precisions this suffices
to overturn the comparison above: against the depthwise $7\times7$ convolution it
replaces, the fused layer trains $1.27$--$1.50\times$ faster in \texttt{fp32} and
$1.40$--$1.76\times$ faster in \texttt{fp16}. It does not win
against a depthwise convolution at its \emph{own} kernel size, which remains
$1.32$--$1.57\times$ faster in \texttt{fp32} and $2.34$--$2.94\times$ in
\texttt{fp16}; that kernel moves $2N$ elements against the fused layer's
$7N$--$8.33N$ and carries no decomposition at all, so the direction is expected.
The speedup over the reference is \emph{smaller} in \texttt{fp16}
than in \texttt{fp32} because half precision halves the bytes moved by both implementations,
and the reference, which moves more of them, benefits more.

\paragraph{Inference.}
\label{sec:results-latency-inference}
\autoref{tab:latency} reports forward-only latency. Against the reference the
picture is unchanged: the fused layer is $3.67$--$4.37\times$ faster in
\texttt{fp32} and $3.38$--$3.60\times$ in \texttt{fp16}, and the reference remains
slower than both plain convolutions in the table, including the depthwise
$7\times7$ it is meant to replace, by $3.79$--$5.28\times$ and
$2.05$--$2.70\times$. The comparison against the baselines, however, does not
reverse as cleanly as over a training step. Against the depthwise $7\times7$
convolution the fused layer wins in \texttt{fp16} ($1.31$--$1.65\times$) but not
in \texttt{fp32}, where it sits at $0.83$--$0.97\times$.

\begin{table}[htpb]
\centering
\caption{Forward latency of every method relative to the reference WTConv ($1.00\times$; higher is faster).}
\label{tab:latency}
\begin{tabular}{llccccc}
\toprule
\multirow{2}{*}{Method} & \multirow{2}{*}{Precision} & \multicolumn{5}{c}{Decomposition levels $L$} \\
\cmidrule(lr){3-7}
 & & 1 & 2 & 3 & 4 & 5 \\
\midrule
Depthwise conv, $k{=}5$ & \texttt{fp32} & 7.35$\times$ & 9.28$\times$ & 9.87$\times$ & 10.10$\times$ & 10.24$\times$ \\
 & \texttt{fp16} & 9.79$\times$ & 11.95$\times$ & 12.54$\times$ & 12.75$\times$ & 12.89$\times$ \\
\addlinespace[2pt]
Depthwise conv, $k{=}7$ & \texttt{fp32} & 3.79$\times$ & 4.79$\times$ & 5.09$\times$ & 5.21$\times$ & 5.28$\times$ \\
 & \texttt{fp16} & 2.05$\times$ & 2.50$\times$ & 2.62$\times$ & 2.66$\times$ & 2.70$\times$ \\
\addlinespace[2pt]
WTConv, reference & \texttt{fp32} & 1.00$\times$ & 1.00$\times$ & 1.00$\times$ & 1.00$\times$ & 1.00$\times$ \\
 & \texttt{fp16} & 1.00$\times$ & 1.00$\times$ & 1.00$\times$ & 1.00$\times$ & 1.00$\times$ \\
\addlinespace[2pt]
WTConv, \textbf{fused} & \texttt{fp32} & 3.67$\times$ & 4.20$\times$ & 4.32$\times$ & 4.35$\times$ & 4.37$\times$ \\
 & \texttt{fp16} & 3.38$\times$ & 3.60$\times$ & 3.60$\times$ & 3.57$\times$ & 3.53$\times$ \\
\addlinespace[2pt]
\bottomrule
\end{tabular}
\end{table}

\subsection{Peak memory}
\label{sec:results-memory}

\paragraph{Training step.} \autoref{tab:memory} reports peak allocated memory over
a training step as a fraction of what the reference WTConv allocates. The fused
layer needs $0.43$--$0.55\times$ the memory of the reference (a reduction by a
factor of $1.83$--$2.31$), and the reduction is essentially flat in $L$ beyond
$L{=}2$, because the dominant saved allocation is the level-1 coefficient tensor:
it holds $N$ elements, three times as many as all deeper levels combined
($\sum_{\ell\ge2}N_\ell\to N/3$). This behavior follows from the mechanism described in \autoref{sec:method}: every per-level subband tensor the reference writes to HBM is also
retained by autograd until the backward pass, whereas the fused layer recomputes
the Haar coefficients on chip. Against the plain convolution, the fused layer's
training-step footprint is $0.87$--$1.02\times$ that of the depthwise $7\times7$
convolution in \texttt{fp32}, and lower still in \texttt{fp16}
($0.76$--$0.89\times$). The fused $L$-level layer therefore has a training footprint
within a few percent of the memory budget of a single plain convolution and falls
below it in half precision.

\begin{table}[htpb]
\centering
\caption{Peak allocated memory over a training step, as a fraction of what the reference WTConv allocates (lower is better).}
\label{tab:memory}
\begin{tabular}{llccccc}
\toprule
\multirow{2}{*}{Method} & \multirow{2}{*}{Precision} & \multicolumn{5}{c}{Decomposition levels $L$} \\
\cmidrule(lr){3-7}
 & & 1 & 2 & 3 & 4 & 5 \\
\midrule
Depthwise conv, $k{=}5$ & \texttt{fp32} & 0.63$\times$ & 0.44$\times$ & 0.44$\times$ & 0.44$\times$ & 0.44$\times$ \\
 & \texttt{fp16} & 0.73$\times$ & 0.51$\times$ & 0.51$\times$ & 0.51$\times$ & 0.51$\times$ \\
\addlinespace[2pt]
Depthwise conv, $k{=}7$ & \texttt{fp32} & 0.63$\times$ & 0.44$\times$ & 0.44$\times$ & 0.44$\times$ & 0.44$\times$ \\
 & \texttt{fp16} & 0.73$\times$ & 0.51$\times$ & 0.51$\times$ & 0.51$\times$ & 0.51$\times$ \\
\addlinespace[2pt]
WTConv, reference & \texttt{fp32} & 1.00$\times$ & 1.00$\times$ & 1.00$\times$ & 1.00$\times$ & 1.00$\times$ \\
 & \texttt{fp16} & 1.00$\times$ & 1.00$\times$ & 1.00$\times$ & 1.00$\times$ & 1.00$\times$ \\
\addlinespace[2pt]
WTConv, \textbf{fused} & \texttt{fp32} & 0.55$\times$ & 0.43$\times$ & 0.44$\times$ & 0.45$\times$ & 0.45$\times$ \\
 & \texttt{fp16} & 0.55$\times$ & 0.44$\times$ & 0.45$\times$ & 0.45$\times$ & 0.45$\times$ \\
\addlinespace[2pt]
\bottomrule
\end{tabular}
\end{table}

\begin{table}[htpb]
\centering
\caption{Peak allocated memory for inference, as a fraction of what the reference WTConv allocates (lower is better).}
\label{tab:memory-inference}
\begin{tabular}{llccccc}
\toprule
\multirow{2}{*}{Method} & \multirow{2}{*}{Precision} & \multicolumn{5}{c}{Decomposition levels $L$} \\
\cmidrule(lr){3-7}
 & & 1 & 2 & 3 & 4 & 5 \\
\midrule
Depthwise conv, $k{=}5$ & \texttt{fp32} & 0.42$\times$ & 0.29$\times$ & 0.29$\times$ & 0.30$\times$ & 0.30$\times$ \\
 & \texttt{fp16} & 0.65$\times$ & 0.45$\times$ & 0.46$\times$ & 0.47$\times$ & 0.47$\times$ \\
\addlinespace[2pt]
Depthwise conv, $k{=}7$ & \texttt{fp32} & 0.56$\times$ & 0.39$\times$ & 0.40$\times$ & 0.40$\times$ & 0.40$\times$ \\
 & \texttt{fp16} & 0.66$\times$ & 0.46$\times$ & 0.47$\times$ & 0.47$\times$ & 0.47$\times$ \\
\addlinespace[2pt]
WTConv, reference & \texttt{fp32} & 1.00$\times$ & 1.00$\times$ & 1.00$\times$ & 1.00$\times$ & 1.00$\times$ \\
 & \texttt{fp16} & 1.00$\times$ & 1.00$\times$ & 1.00$\times$ & 1.00$\times$ & 1.00$\times$ \\
\addlinespace[2pt]
WTConv, \textbf{fused} & \texttt{fp32} & 0.65$\times$ & 0.50$\times$ & 0.52$\times$ & 0.53$\times$ & 0.53$\times$ \\
 & \texttt{fp16} & 0.65$\times$ & 0.51$\times$ & 0.53$\times$ & 0.54$\times$ & 0.54$\times$ \\
\addlinespace[2pt]
\bottomrule
\end{tabular}
\end{table}

\paragraph{Inference.}
\label{sec:results-memory-inference}
\autoref{tab:memory-inference} is the forward-only counterpart, with no autograd
tape. The fused/reference footprint ratio is $0.50$--$0.65\times$ during inference,
versus $0.43$--$0.55\times$ during training. The smaller reduction at inference is
expected because a large part of what fusion removes is tape-retained subband
tensors that inference never allocates in the first place; what remains is the
traffic-side reduction of \autoref{eq:qfused}. The plain convolutions are
correspondingly harder to match, since without a tape their footprint is close to
the compulsory input plus output: the fused layer sits at $1.15$--$1.34\times$ the
depthwise $7\times7$ convolution's footprint in \texttt{fp32} and
$0.99$--$1.15\times$ in \texttt{fp16}, the residual being the filtered subband
coefficients it must materialize between the analysis and synthesis passes.

\subsection{Generality across kernel size}
\label{sec:results-kernel-size}

Every measurement above runs at $k{=}5$, the kernel size WTConvNeXt deploys. The modeled
tensor-materialization traffic, however, is independent of $k$: \autoref{eq:qref} and
\autoref{eq:qfused} are both independent of $k$, which sets how much arithmetic each resident
element receives rather than how many elements cross the HBM boundary. \autoref{tab:k3} tests
that claim by repeating the sweep of \autoref{sec:setup} at $k{=}3$.

\begin{table}[htpb]
\centering
\caption{The fused layer compared with the reference WTConv, both at $k{=}3$. Latency is reported as speedup relative to the reference ($1.00\times$; higher is faster), and peak allocated memory as a fraction of the reference ($1.00\times$; lower is better).}
\label{tab:k3}
\begin{tabular}{llccccc}
\toprule
\multirow{2}{*}{Quantity} & \multirow{2}{*}{Precision} & \multicolumn{5}{c}{Decomposition levels $L$} \\
\cmidrule(lr){3-7}
 & & 1 & 2 & 3 & 4 & 5 \\
\midrule
Training-step speedup & \texttt{fp32} & 3.86$\times$ & 4.50$\times$ & 4.63$\times$ & 4.67$\times$ & 4.68$\times$ \\
 & \texttt{fp16} & 3.31$\times$ & 3.82$\times$ & 3.93$\times$ & 3.94$\times$ & 3.94$\times$ \\
\addlinespace[2pt]
Inference speedup & \texttt{fp32} & 4.48$\times$ & 5.06$\times$ & 5.24$\times$ & 5.28$\times$ & 5.32$\times$ \\
 & \texttt{fp16} & 4.40$\times$ & 4.81$\times$ & 4.84$\times$ & 4.82$\times$ & 4.77$\times$ \\
\addlinespace[2pt]
Peak memory, training & \texttt{fp32} & 0.54$\times$ & 0.43$\times$ & 0.44$\times$ & 0.44$\times$ & 0.44$\times$ \\
 & \texttt{fp16} & 0.54$\times$ & 0.43$\times$ & 0.44$\times$ & 0.44$\times$ & 0.44$\times$ \\
\addlinespace[2pt]
Peak memory, inference & \texttt{fp32} & 0.64$\times$ & 0.49$\times$ & 0.52$\times$ & 0.52$\times$ & 0.52$\times$ \\
 & \texttt{fp16} & 0.64$\times$ & 0.49$\times$ & 0.52$\times$ & 0.52$\times$ & 0.52$\times$ \\
\addlinespace[2pt]
\bottomrule
\end{tabular}
\end{table}

The speedup over the reference is larger at $k{=}3$ than at $k{=}5$ in every cell of the table. A
training step is $3.86$--$4.68\times$ faster than the reference in \texttt{fp32} and
$3.31$--$3.94\times$ in \texttt{fp16}, against $3.71$--$4.35\times$ and $2.68$--$3.09\times$ at
$k{=}5$; inference is $4.48$--$5.32\times$ and $4.40$--$4.84\times$, against $3.67$--$4.37\times$
and $3.38$--$3.60\times$. \autoref{sec:memorybound} predicts this direction. Shrinking the kernel
removes arithmetic while leaving \autoref{eq:qref} untouched, so the reference falls further into
the memory-bound regime.

\subsection{End-to-end networks}
\label{sec:results-endtoend}

To evaluate the operator at architectural scale, we benchmark ConvNeXt-T and WTConvNeXt-T \citep{finder2024wavelet}, which replaces ConvNeXt-T's depthwise convolutions with WTConv ($k{=}5$, $L{\in}\{5,4,3,2\}$). We use a batch size of 64, $224\times224$ inputs, and \texttt{fp32}; \autoref{tab:endtoend} reports the results. Because the reformulation preserves the mathematical operator up to floating-point evaluation order, it does not alter the model architecture or learned parameters.

\begin{table}[htpb]
\centering
\caption{End-to-end network throughput and peak allocated GPU memory, for a
forward pass and for a full training step (forward and backward).}
\label{tab:endtoend}
\setlength{\tabcolsep}{4pt}
\begin{tabular}{llcccc}
\toprule
 & & ConvNeXt-T & \multicolumn{2}{c}{WTConvNeXt-T} & Fused / \\
\cmidrule(lr){4-5}
 & & (depthwise) & reference & fused & reference \\
\midrule
\multirow{2}{*}{Inference}
 & Throughput (img/s) $\uparrow$ & $1394.9$ & $420.7$ ($0.30\times$) & $988.5$ ($0.71\times$) & $2.35\times$ \\
 & Peak memory (MiB) $\downarrow$ & $596.8$ & $887.4$ ($1.49\times$) & $750.8$ ($1.26\times$) & $0.85\times$ \\
\addlinespace[3pt]
\multirow{2}{*}{Training step}
 & Throughput (img/s) $\uparrow$ & $247.7$ & $166.3$ ($0.67\times$) & $263.0$ ($1.06\times$) & $1.58\times$ \\
 & Peak memory (MiB) $\downarrow$ & $6933.2$ & $9685.6$ ($1.40\times$) & $7618.4$ ($1.10\times$) & $0.79\times$ \\
\bottomrule
\end{tabular}
\end{table}

The reference WTConv imposes a severe bottleneck, restricting the network to $30\%$ of ConvNeXt-T's inference throughput and $67\%$ of its training throughput. Our fused implementation effectively mitigates this, achieving $2.35\times$ and $1.58\times$ the reference's inference and training throughputs, respectively. Crucially, WTConvNeXt-T equipped with the fused layer trains $1.06\times$ faster than the baseline ConvNeXt-T, although inference throughput reaches only $0.71\times$ of the baseline.

The fused formulation similarly reduces peak memory, cutting the reference WTConvNeXt-T's training and inference footprints to $0.79\times$ and $0.85\times$. This limits the network's peak memory to $1.10\times$ that of ConvNeXt-T during training and $1.26\times$ during inference.

\subsection{Ablation of the reformulations}
\label{sec:results-ablation}

We evaluate the three reformulations cumulatively, in their order in \autoref{sec:method}. The first
variant fuses Haar analysis and depthwise convolution at each level. The second also
replaces the sequential synthesis loop with the bit-indexed pass of \autoref{prop:bitindexed}. The
full method further folds the base-path scale and fuses its addition into the final store. Each
variant otherwise follows the reference \texttt{WTConv2d} implementation and is verified against
its outputs and parameter gradients (\autoref{app:extra}). We measure full training steps at
$k{=}5$ in \texttt{fp32}, using the $(C,H)$ sweep of \autoref{sec:setup}; \autoref{tab:ablation}
reports latency and peak memory.

\begin{table}[htpb]
\centering
\caption{Cumulative ablation of the reformulations over a training step at $k{=}5$ in \texttt{fp32}, relative to the reference WTConv ($1.00\times$). Latency is reported as speedup (higher is faster), and peak allocated memory as a fraction of the reference (lower is better).}
\label{tab:ablation}
\begin{tabular}{lccccc}
\toprule
\multirow{2}{*}{Method} & \multicolumn{5}{c}{Decomposition levels $L$} \\
\cmidrule(lr){2-6}
 & 1 & 2 & 3 & 4 & 5 \\
\midrule
\multicolumn{6}{l}{\emph{Training-step speedup}} \\
WTConv, reference & 1.00$\times$ & 1.00$\times$ & 1.00$\times$ & 1.00$\times$ & 1.00$\times$ \\
+ Fused Haar analysis (\autoref{sec:analysis}) & 1.35$\times$ & 1.44$\times$ & 1.46$\times$ & 1.45$\times$ & 1.44$\times$ \\
+ Collapsed synthesis (\autoref{sec:synthesis}) & 2.54$\times$ & 2.95$\times$ & 3.02$\times$ & 3.04$\times$ & 3.05$\times$ \\
+ Scale folding (\autoref{sec:scale_folding}) = Full (\textbf{Ours}) & 3.71$\times$ & 4.23$\times$ & 4.34$\times$ & 4.35$\times$ & 4.33$\times$ \\
\midrule
\multicolumn{6}{l}{\emph{Peak memory, training}} \\
WTConv, reference & 1.00$\times$ & 1.00$\times$ & 1.00$\times$ & 1.00$\times$ & 1.00$\times$ \\
+ Fused Haar analysis (\autoref{sec:analysis}) & 0.79$\times$ & 0.60$\times$ & 0.59$\times$ & 0.59$\times$ & 0.59$\times$ \\
+ Collapsed synthesis (\autoref{sec:synthesis}) & 0.65$\times$ & 0.49$\times$ & 0.50$\times$ & 0.50$\times$ & 0.50$\times$ \\
+ Scale folding (\autoref{sec:scale_folding}) = Full (\textbf{Ours}) & 0.55$\times$ & 0.43$\times$ & 0.44$\times$ & 0.45$\times$ & 0.45$\times$ \\
\bottomrule
\end{tabular}
\end{table}

Each reformulation improves both latency and peak memory at every $L$. Fusing Haar analysis alone
gives $1.35$--$1.46\times$ and lowers the footprint to $0.59$--$0.79\times$. Collapsing the
synthesis cascade contributes the largest single gain, a further $1.88$--$2.12\times$, bringing the
speedup to $2.54$--$3.05\times$ and memory to $0.49$--$0.65\times$. Folding the scales and fusing
the base-path addition adds a further $1.42$--$1.46\times$, reaching the full $3.71$--$4.35\times$
and $0.43$--$0.55\times$ reported in \autoref{sec:results-latency-train} and
\autoref{sec:results-memory}.

\subsection{Generality across hardware}
\label{sec:results-hardware}

All preceding layer measurements use an RTX~A6000 (Ampere, sm\_86). To assess device dependence,
\autoref{tab:hardware} repeats the $k{=}5$ sweep on an NVIDIA RTX~PRO~6000 Blackwell Max-Q
Workstation Edition (sm\_120).

\begin{table}[htpb]
\centering
\caption{The fused layer compared with the reference WTConv, both measured on the second GPU, $k{=}5$. Latency is reported as speedup relative to the reference ($1.00\times$; higher is faster), and peak allocated memory as a fraction of the reference ($1.00\times$; lower is better).}
\label{tab:hardware}
\begin{tabular}{llccccc}
\toprule
\multirow{2}{*}{Quantity} & \multirow{2}{*}{Precision} & \multicolumn{5}{c}{Decomposition levels $L$} \\
\cmidrule(lr){3-7}
 & & 1 & 2 & 3 & 4 & 5 \\
\midrule
Training-step speedup & \texttt{fp32} & 2.49$\times$ & 2.53$\times$ & 2.79$\times$ & 2.46$\times$ & 2.48$\times$ \\
 & \texttt{fp16} & 1.95$\times$ & 2.07$\times$ & 2.07$\times$ & 2.41$\times$ & 2.42$\times$ \\
\addlinespace[2pt]
Inference speedup & \texttt{fp32} & 2.87$\times$ & 3.01$\times$ & 3.03$\times$ & 3.02$\times$ & 3.05$\times$ \\
 & \texttt{fp16} & 2.92$\times$ & 3.07$\times$ & 3.10$\times$ & 3.10$\times$ & 3.07$\times$ \\
\addlinespace[2pt]
Peak memory, training & \texttt{fp32} & 0.55$\times$ & 0.43$\times$ & 0.44$\times$ & 0.45$\times$ & 0.45$\times$ \\
 & \texttt{fp16} & 0.49$\times$ & 0.40$\times$ & 0.41$\times$ & 0.41$\times$ & 0.41$\times$ \\
\addlinespace[2pt]
Peak memory, inference & \texttt{fp32} & 0.65$\times$ & 0.50$\times$ & 0.52$\times$ & 0.53$\times$ & 0.53$\times$ \\
 & \texttt{fp16} & 0.65$\times$ & 0.51$\times$ & 0.53$\times$ & 0.54$\times$ & 0.54$\times$ \\
\addlinespace[2pt]
\bottomrule
\end{tabular}
\end{table}

The fused implementation remains faster for all $L$ and both precisions. Relative to the
reference, training is $2.46$--$2.79\times$ faster in \texttt{fp32} and $1.95$--$2.42\times$ faster
in \texttt{fp16}; inference is $2.87$--$3.05\times$ and $2.92$--$3.10\times$ faster, respectively.
The realized speedup is device-dependent even though both implementations remain well below the device’s roofline ridge point. Differences in achieved bandwidth, kernel dispatch, launch overhead, cache behavior, and kernel utilization are not captured by the tensor-traffic model. Peak-memory
reductions are stable across devices, since they are governed mainly by tensor materialization
rather than GPU throughput, matching the A6000 trends at $0.40$--$0.55\times$ for training and
$0.50$--$0.65\times$ for inference.


\section{Limitations}

Our reformulation relies on properties specific to the Haar wavelet, including its low-arithmetic-cost, symmetric, and self-inverse transform structure. In particular, the fused analysis and bit-indexed closed-form synthesis do not directly generalize to wavelet families with longer filters or more complex reconstruction rules. Extending the approach beyond Haar would therefore require new transform-specific formulations.


\section{Conclusion}

WTConv's performance bottleneck is not arithmetic but data movement. Although the operator performs only moderately more computation than the depthwise convolution it replaces, its reference implementation repeatedly materializes intermediate wavelet coefficients and reconstructions in HBM, leaving it deeply memory-bound. By making this I/O cost explicit, we derived an algebraically equivalent formulation that keeps Haar analysis on chip, collapses the multi-level synthesis recursion into a single bit-indexed pass, and folds learned scales into the convolution weights.

These reformulations reduce modeled HBM traffic by approximately $2.55\times$ and translate directly into substantial practical gains. Across the evaluated configurations, the fused implementation is $3.71$--$4.35\times$ faster than the reference in \texttt{fp32} and $2.68$--$3.09\times$ faster in \texttt{fp16} over a full training step, while reducing peak memory by a factor of $1.83$--$2.31$. More importantly, the optimization changes the practical trade-off that motivates WTConv: the fused layer trains $1.27$--$1.50\times$ faster in \texttt{fp32} and $1.40$--$1.76\times$ faster in \texttt{fp16} than the depthwise $7\times7$ convolution it is intended to replace.

The broader lesson is that favorable FLOP counts and parameter scaling do not by themselves imply an efficient operator. For structured, multi-stage layers built from inexpensive transforms, intermediate tensor materialization can dominate execution cost. In such settings, I/O-aware algebraic reformulation is not merely an implementation optimization; it can determine whether the theoretical advantages of an operator translate into practical gains.

\subsubsection*{Broader Impact Statement}
This work reduces the time and memory required to evaluate an existing operator without changing its function; the primary effect is lower computational cost for practitioners already using WTConv. We do not identify application-specific ethical risks beyond the general risks associated with making computer-vision systems more computationally efficient.

\section*{Acknowledgments}

This work was partially supported by the Office of Naval Research Global under Grant No.~N629092312104.

\clearpage
\bibliography{main}
\bibliographystyle{tmlr}

\appendix

\clearpage
\section{Why the bit-indexed Haar synthesis formula is correct}
\label{app:bitindexed-proof}

We first recall what a single Haar synthesis step does.  Given the four
coefficients $(c_{\mathrm{LL}},c_{\mathrm{LH}},c_{\mathrm{HL}},
c_{\mathrm{HH}})$ of one $2\times2$ block, the value reconstructed at
within-block position $(r_y,r_x)\in\{0,1\}^2$ is
\begin{equation}
\frac{1}{2}\left(
  c_{\mathrm{LL}}
  + (-1)^{r_y}c_{\mathrm{LH}}
  + (-1)^{r_x}c_{\mathrm{HL}}
  + (-1)^{r_y+r_x}c_{\mathrm{HH}}
\right).
\label{eq:haar-synthesis-local}
\end{equation}
This is simply the corresponding row of the Haar synthesis matrix
$\mH^{\!\top}=\mH$.  The sign pattern is easier to see explicitly:
\begin{equation}
\begin{array}{c|c|ccc}
(r_y,r_x) & \text{position} & c_{\mathrm{LH}} & c_{\mathrm{HL}}
                                      & c_{\mathrm{HH}} \\
\hline
(0,0) & \text{top left}     & + & + & + \\
(0,1) & \text{top right}    & + & - & - \\
(1,0) & \text{bottom left}  & - & + & - \\
(1,1) & \text{bottom right} & - & - & +
\end{array}
\label{eq:haar-sign-table}
\end{equation}

We now apply this observation to level $\ell$.  A coefficient at this level
reconstructs a $2^\ell\times2^\ell$ block of the final image.  The coefficient
tuple that affects output pixel $(y,x)$ is therefore the one at coarse-grid
location
\begin{equation}
\left(\left\lfloor\frac{y}{2^\ell}\right\rfloor,
      \left\lfloor\frac{x}{2^\ell}\right\rfloor\right).
\label{eq:level-coefficient-address}
\end{equation}
Within that block, the bit
\begin{equation}
b_y^{(\ell)}
= \left\lfloor\frac{y}{2^{\ell-1}}\right\rfloor\bmod2
\end{equation}
tells whether $y$ lies in the top half ($b_y^{(\ell)}=0$) or bottom half
($b_y^{(\ell)}=1$).  Similarly,
$b_x^{(\ell)}=\lfloor x/2^{\ell-1}\rfloor\bmod2$ selects the left or right
half.  Hence the signs in \autoref{eq:haar-synthesis-local} are exactly
\begin{equation}
(-1)^{b_y^{(\ell)}}=s_y^{(\ell)}, \qquad
(-1)^{b_x^{(\ell)}}=s_x^{(\ell)}, \qquad
(-1)^{b_y^{(\ell)}+b_x^{(\ell)}}
=s_y^{(\ell)}s_x^{(\ell)}.
\label{eq:level-signs}
\end{equation}
In other words, dividing by $2^{\ell-1}$ discards the lower coordinate bits,
and reducing modulo two extracts precisely the bit that selects the quadrant
in \autoref{eq:haar-sign-table}.

It remains to explain the factor $2^{-\ell}$.  The four level-$\ell$ bands
first undergo the synthesis step in \autoref{eq:haar-synthesis-local}, which
contributes a factor $1/2$.  Their result then travels through the LL input of
the remaining $\ell-1$ finer synthesis steps.  In an LL-only step, all four
children receive the parent value with positive sign and another factor
$1/2$.  There are therefore $\ell$ factors of $1/2$ in total:
\begin{equation}
\underbrace{\frac12\cdots\frac12}_{\ell\ \mathrm{times}}=2^{-\ell}.
\label{eq:level-normalization}
\end{equation}

For example, a level-$1$ contribution uses the lowest bits
$(y\bmod2,x\bmod2)$ and is weighted by $1/2$.  A level-$2$ contribution uses
the next bits $(\lfloor y/2\rfloor\bmod2,
\lfloor x/2\rfloor\bmod2)$ and is weighted by $1/4$: its bands are combined
once, and the result passes through one additional LL-only synthesis step.

Finally, Haar synthesis and the recursive LL additions are linear.  We may
therefore compute the contribution of each level independently and add the
results.  Substituting the address in
\autoref{eq:level-coefficient-address}, the signs in
\autoref{eq:level-signs}, and the normalization in
\autoref{eq:level-normalization}, then summing over
$\ell=1,\ldots,L$, gives \autoref{eq:synthesis_closed_form}.

\clearpage

\section{Empirical validation of the I/O model}
\label{app:traffic}

\autoref{eq:qref} and \autoref{eq:qfused} predict HBM traffic by
assuming that each implementation stage reads its input once and writes its
output once. We evaluate this assumption using hardware performance counters.
The fused implementation closely follows the model, whereas the reference
matches it only when the library-selected kernels follow the assumed access
pattern.

\paragraph{Method.}
We profile one forward pass with Nsight Compute on an RTX~A6000. We sum
\texttt{dram\_\_bytes\_read.sum} and
\texttt{dram\_\_bytes\_write.sum} over all kernels and divide by the element
size and $N$, yielding the normalized traffic reported in
\autoref{tab:traffic-measured}. We warm up each layer before profiling to stabilize
cuDNN algorithm selection and exclude allocator and extension-initialization
overheads. Unless noted otherwise, we use \texttt{fp32}, $B{=}8$, and $k{=}5$.

\subsection{Agreement under the modeled access pattern}
\label{app:traffic-good}

For $C{=}128$ and $H{=}W{=}128$ ($N=16.8$M), the selected kernels follow the
access pattern assumed in \autoref{sec:memorybound}. The measured and predicted
traffic are compared in \autoref{tab:traffic-measured}.

\begin{table}[h]
\centering
\caption{Predicted and measured HBM traffic, normalized by
$N=B\cdot C\cdot H\cdot W$, for $B{=}8$, $C{=}128$, $H{=}W{=}128$, $k{=}5$,
and \texttt{fp32}.}
\label{tab:traffic-measured}
\begin{tabular}{llccccc}
\toprule
& & \multicolumn{5}{c}{Decomposition levels $L$}\\
\cmidrule(lr){3-7}
& & 1 & 2 & 3 & 4 & 5 \\
\midrule
\multirow{3}{*}{Reference, $\Qio_{\mathrm{ref}}$}
  & \autoref{eq:qref}   & $17.75N$ & $20.44N$ & $21.11N$ & $21.28N$ & $21.32N$ \\
  & measured            & $17.52N$ & $20.83N$ & $21.79N$ & $21.97N$ & $22.13N$ \\
  & measured/predicted  & $0.99\times$ & $1.02\times$ & $1.03\times$ & $1.03\times$ & $1.04\times$ \\
\addlinespace[3pt]
\multirow{3}{*}{Fused, $\Qio_{\mathrm{fused}}$}
  & \autoref{eq:qfused} & $7.00N$  & $8.00N$  & $8.25N$  & $8.31N$  & $8.33N$ \\
  & measured            & $6.96N$  & $7.94N$  & $8.18N$  & $8.24N$  & $8.25N$ \\
  & measured/predicted  & $0.99\times$ & $0.99\times$ & $0.99\times$ & $0.99\times$ & $0.99\times$ \\
\addlinespace[3pt]
\multirow{2}{*}{Traffic reduction}
  & \autoref{eq:qratio} & $2.54\times$ & $2.55\times$ & $2.56\times$ & $2.56\times$ & $2.56\times$ \\
  & measured            & $2.52\times$ & $2.62\times$ & $2.66\times$ & $2.67\times$ & $2.68\times$ \\
\bottomrule
\end{tabular}
\end{table}

Across all levels, the fused measurements are within $1\%$ of
\autoref{eq:qfused}, and the reference measurements are within $4\%$ of
\autoref{eq:qref}. Consequently, the measured traffic reduction of
$2.52$--$2.68\times$ closely agrees with the predicted $2.54$--$2.56\times$.

\subsection{Effect of library kernel selection}
\label{app:traffic-bad}

The reference implementation depends on library kernel selection. Changing only
the shape to $C{=}64$ and $H{=}W{=}256$ causes its measured traffic to exceed
\autoref{eq:qref} by up to $39\%$ (\autoref{tab:traffic-dispatch}), while the
fused implementation remains within $1\%$ of \autoref{eq:qfused}.

\begin{table}[h]
\centering
\caption{Measured HBM traffic and its ratio to the prediction for $B{=}8$,
$C{=}64$, $H{=}W{=}256$, $k{=}5$, and \texttt{fp32}.}
\label{tab:traffic-dispatch}
\begin{tabular}{llccccc}
\toprule
& & \multicolumn{5}{c}{Decomposition levels $L$}\\
\cmidrule(lr){3-7}
& & 1 & 2 & 3 & 4 & 5 \\
\midrule
\multirow{2}{*}{Reference}
  & measured               & $17.98N$ & $27.97N$ & $28.80N$ & $28.82N$ & $29.69N$ \\
  & vs.\ \autoref{eq:qref} & $1.01\times$ & $1.37\times$ & $1.36\times$ & $1.35\times$ & $1.39\times$ \\
\addlinespace[3pt]
\multirow{2}{*}{Fused}
  & measured                 & $6.98N$ & $7.95N$ & $8.19N$ & $8.22N$ & $8.23N$ \\
  & vs.\ \autoref{eq:qfused} & $1.00\times$ & $0.99\times$ & $0.99\times$ & $0.99\times$ & $0.99\times$ \\
\bottomrule
\end{tabular}
\end{table}

The excess originates from the grouped \texttt{conv\_transpose2d} used for Haar
synthesis. From $L{=}2$ onward, cuDNN selects
\texttt{dgrad2d\_alg1\_1} for the shallow levels. At $L{=}2$, this kernel reads
$0.54N$ elements but writes $6.80N$, because it repeatedly accumulates into an
output tensor that exceeds the L2 capacity. The discontinuity between $L{=}1$
and $L{=}2$, and its dependence on tensor shape, identify kernel dispatch rather
than the WTConv algorithm as the source of the additional traffic.

Thus, \autoref{eq:qref} should be interpreted as the algorithmic traffic of the
reference implementation, and as a lower bound when library kernels perform
additional accesses. For this shape, the measured reduction is
$2.58$--$3.61\times$, compared with the predicted $2.54$--$2.56\times$. The
model therefore gives a conservative estimate of the practical reduction. In
contrast, the fused implementation uses kernels with explicit access patterns
and remains close to the prediction across the evaluated configurations.

\subsection{Dependence on kernel size and tensor shape}
\label{app:traffic-invariance}

Both models are independent of kernel size $k$ and, after normalization by $N$,
of tensor shape. \autoref{tab:traffic-invariance} evaluates these predictions.

\begin{table}[h]
\centering
\caption{Measured HBM traffic at $L{=}3$ in \texttt{fp32} across tensor shapes
(left) and kernel sizes (right). The predicted reference and fused traffic are
$21.11N$ and $8.25N$, respectively.}
\label{tab:traffic-invariance}
\begin{tabular}{lcccccc}
\toprule
& \multicolumn{3}{c}{Tensor shape, $k{=}5$} & \multicolumn{3}{c}{Kernel size, $C{=}64$, $H{=}W{=}256$}\\
\cmidrule(lr){2-4}\cmidrule(lr){5-7}
& $C{=}32$  & $C{=}64$  & $C{=}128$ & \multirow{2}{*}{$k{=}3$} & \multirow{2}{*}{$k{=}5$} & \multirow{2}{*}{$k{=}7$} \\
& $H{=}512$ & $H{=}256$ & $H{=}128$ & & & \\
\midrule
Reference & $26.97N$ & $29.88N$ & $21.78N$ & $29.61N$ & $29.88N$ & $28.59N$ \\
Fused     & $8.60N$  & $8.17N$  & $8.18N$  & $8.19N$  & $8.17N$  & $8.19N$ \\
\bottomrule
\end{tabular}
\end{table}

Traffic shows no systematic dependence on $k$: over $k\in\{3,5,7\}$, the fused
measurements vary by $0.02N$ and the reference measurements by approximately
$1N$. Across shapes whose element counts differ by $4\times$, fused traffic
remains within $5\%$ of the prediction. The larger variation in the reference
measurements ($21.78N$--$29.88N$) is consistent with the dispatch effect
described in \autoref{app:traffic-bad}.

\clearpage

\section{Supplementary measurements}
\label{app:extra}

\subsection{Absolute layer timings}

\autoref{tab:absolute-k5} and \autoref{tab:absolute-k3} report the absolute
latencies underlying the ratios in \autoref{tab:latency-train},
\autoref{tab:latency}, and \autoref{tab:k3}. Each entry is the geometric mean
over the nine $(C,H)$ configurations in \autoref{sec:setup} of the mean latency
from 50 timed iterations. Consequently, before the displayed values are
rounded, dividing a reference entry by another entry reproduces the
corresponding geometric-mean speedup in the main paper. Plain convolutions do
not depend on $L$, so their measured latency is repeated across the five
columns.

\begin{table}[htpb]
\centering
\small
\caption{Absolute latency in milliseconds for the $k{=}5$ sweep. WTConv uses
$k{=}5$ throughout. The depthwise $k{=}5$ row is the matched-kernel baseline,
and the depthwise $k{=}7$ row is the ConvNeXt drop-in baseline.}
\label{tab:absolute-k5}
\begin{tabular}{llccccc}
\toprule
\multirow{2}{*}{Method} & \multirow{2}{*}{Precision} & \multicolumn{5}{c}{Decomposition levels $L$} \\
\cmidrule(lr){3-7}
 & & 1 & 2 & 3 & 4 & 5 \\
\midrule
\multicolumn{7}{l}{\textit{Forward pass}} \\
\addlinespace[2pt]
Depthwise conv, $k{=}5$ & \texttt{fp32} & 1.154 & 1.154 & 1.154 & 1.154 & 1.154 \\
 & \texttt{fp16} & 0.479 & 0.479 & 0.479 & 0.479 & 0.479 \\
\addlinespace[2pt]
Depthwise conv, $k{=}7$ & \texttt{fp32} & 2.237 & 2.237 & 2.237 & 2.237 & 2.237 \\
 & \texttt{fp16} & 2.291 & 2.291 & 2.291 & 2.291 & 2.291 \\
\addlinespace[2pt]
WTConv, reference & \texttt{fp32} & 8.487 & 10.710 & 11.387 & 11.652 & 11.815 \\
 & \texttt{fp16} & 4.689 & 5.723 & 6.004 & 6.104 & 6.174 \\
\addlinespace[2pt]
WTConv, \textbf{fused} & \texttt{fp32} & 2.312 & 2.552 & 2.634 & 2.681 & 2.703 \\
 & \texttt{fp16} & 1.389 & 1.592 & 1.668 & 1.710 & 1.748 \\
\midrule
\multicolumn{7}{l}{\textit{Training step (forward and backward)}} \\
\addlinespace[2pt]
Depthwise conv, $k{=}5$ & \texttt{fp32} & 5.913 & 5.913 & 5.913 & 5.913 & 5.913 \\
 & \texttt{fp16} & 2.289 & 2.289 & 2.289 & 2.289 & 2.289 \\
\addlinespace[2pt]
Depthwise conv, $k{=}7$ & \texttt{fp32} & 11.737 & 11.737 & 11.737 & 11.737 & 11.737 \\
 & \texttt{fp16} & 9.417 & 9.417 & 9.417 & 9.417 & 9.417 \\
\addlinespace[2pt]
WTConv, reference & \texttt{fp32} & 28.901 & 36.738 & 39.036 & 39.854 & 40.163 \\
 & \texttt{fp16} & 14.362 & 18.592 & 19.842 & 20.366 & 20.758 \\
\addlinespace[2pt]
WTConv, \textbf{fused} & \texttt{fp32} & 7.800 & 8.693 & 8.995 & 9.154 & 9.277 \\
 & \texttt{fp16} & 5.365 & 6.148 & 6.442 & 6.600 & 6.733 \\
\bottomrule
\end{tabular}
\end{table}

\begin{table}[p]
\centering
\small
\caption{Absolute latency in milliseconds for the reference and fused WTConv
implementations in the $k{=}3$ sweep.}
\label{tab:absolute-k3}
\begin{tabular}{llccccc}
\toprule
\multirow{2}{*}{Method} & \multirow{2}{*}{Precision} & \multicolumn{5}{c}{Decomposition levels $L$} \\
\cmidrule(lr){3-7}
 & & 1 & 2 & 3 & 4 & 5 \\
\midrule
\multicolumn{7}{l}{\textit{Forward pass}} \\
\addlinespace[2pt]
WTConv, reference & \texttt{fp32} & 7.474 & 9.506 & 10.198 & 10.449 & 10.651 \\
 & \texttt{fp16} & 4.529 & 5.552 & 5.834 & 5.940 & 5.999 \\
\addlinespace[2pt]
WTConv, \textbf{fused} & \texttt{fp32} & 1.667 & 1.880 & 1.945 & 1.979 & 2.001 \\
 & \texttt{fp16} & 1.029 & 1.155 & 1.205 & 1.234 & 1.258 \\
\midrule
\multicolumn{7}{l}{\textit{Training step (forward and backward)}} \\
\addlinespace[2pt]
WTConv, reference & \texttt{fp32} & 23.462 & 31.003 & 33.036 & 33.845 & 34.315 \\
 & \texttt{fp16} & 12.955 & 17.093 & 18.310 & 18.821 & 19.186 \\
\addlinespace[2pt]
WTConv, \textbf{fused} & \texttt{fp32} & 6.081 & 6.891 & 7.135 & 7.248 & 7.332 \\
 & \texttt{fp16} & 3.908 & 4.470 & 4.665 & 4.773 & 4.870 \\
\bottomrule
\end{tabular}
\end{table}

\clearpage
\subsection{Numerical agreement}
\autoref{tab:correctness} quantifies the numerical agreement referenced in
\autoref{sec:setup}. 
\begin{table}[htbp]
\centering
\caption{Numerical agreement with the reference implementation. Entries are maximum absolute deviation over the whole tensor, at $L$ decomposition levels, for the forward output and layer gradient. The fused kernels compute the same multilinear map with a different association order, so the residual is floating-point reassociation error alone.}
\label{tab:correctness}
\begin{tabular}{llccc}
\toprule
Backend & dtype & $L$ & forward & gradients \\
\midrule
Fused (CUDA) & \texttt{fp32} & 1 & 2.4e-07 & 2.4e-07 \\
Fused (CUDA) & \texttt{fp32} & 2 & 2.4e-07 & 2.4e-07 \\
Fused (CUDA) & \texttt{fp32} & 3 & 2.4e-07 & 2.4e-07 \\
Fused (CUDA) & \texttt{fp32} & 4 & 2.4e-07 & 2.4e-07 \\
Fused (CUDA) & \texttt{fp32} & 5 & 2.4e-07 & 2.4e-07 \\
\addlinespace[2pt]
Fused (CUDA) & \texttt{fp16} & 1 & 2.0e-03 & 9.8e-04 \\
Fused (CUDA) & \texttt{fp16} & 2 & 2.0e-03 & 2.0e-03 \\
Fused (CUDA) & \texttt{fp16} & 3 & 2.0e-03 & 2.0e-03 \\
Fused (CUDA) & \texttt{fp16} & 4 & 2.0e-03 & 2.0e-03 \\
Fused (CUDA) & \texttt{fp16} & 5 & 2.0e-03 & 2.0e-03 \\
\bottomrule
\end{tabular}
\end{table}

\end{document}